\documentclass{article}

\PassOptionsToPackage{numbers, compress}{natbib}
\usepackage[utf8]{inputenc} 
\usepackage[T1]{fontenc}    
\usepackage{hyperref}       
\usepackage{url}            
\usepackage{booktabs}       
\usepackage{amsfonts}       
\usepackage{nicefrac}       
\usepackage{microtype}      
\usepackage{xcolor}         
\usepackage{amsthm}
\usepackage{amsmath}
\usepackage{mathtools}
\usepackage{subfig}
\usepackage{nicefrac}
\usepackage{cleveref}
\usepackage{xcolor}
\usepackage{bm}
\usepackage{tikz}
\usetikzlibrary{matrix,positioning}
\usepackage{enumerate}
\usepackage{multicol}
\usepackage{wrapfig}
\usepackage{subcaption} 

\usepackage{algpseudocode}
\usepackage[preprint]{neurips_2025}

\usepackage[utf8]{inputenc} 
\usepackage[T1]{fontenc}    
\usepackage{hyperref}       
\usepackage{url}            
\usepackage{booktabs}       
\usepackage{amsfonts}       
\usepackage{nicefrac}       
\usepackage{microtype}      
\usepackage{xcolor}         
\usepackage{wrapfig}

\title{Layer Specialization Underlying Compositional Reasoning in Transformers}

\author{
  Jing Liu \\
  ENS, Université PSL, EHESS, CNRS \\
  Paris, France \\
  \texttt{jing.liu@psl.eu} \\
}

\begin{document}

\maketitle

\begin{abstract}
Transformers exhibit compositional reasoning on sequences not observed during training, a capability often attributed to in-context learning (ICL) and skill composition. We investigate this phenomenon using the Random Hierarchy Model (RHM), a probabilistic context-free grammar that generates sequences through recursive rule application. Models are trained on subsets of sequences and evaluated across four generalization conditions: memorization, in-distribution generalization, out-of-distribution generalization with the same rules, and cross-layer transfer. Behaviorally, performance improves systematically with task complexity and the number of in-context examples, with out-of-distribution tasks requiring substantially more examples than in-distribution scenarios. Mechanistically, we identify a progressive emergence of layer specialization during training that correlates with generalization performance. Principal component analysis and attention pattern clustering reveal that transformers develop structured, hierarchically organized representations in specialized layers. These results demonstrate that transformers develop modular, interpretable mechanisms supporting compositional reasoning, linking internal algorithmic structure to observed behavioral capabilities.
\end{abstract}

\section{Introduction}

Large language models have demonstrated remarkable abilities to perform complex tasks through in-context learning (ICL), adapting to new challenges from just a few demonstrations without parameter updates \citep{brown2020language}. A key aspect of this capability is \emph{compositional reasoning}—the ability to understand and generate novel combinations of learned elements by flexibly recombining previously acquired skills \citep{lake2018generalization}. Understanding how neural networks acquire and deploy compositional abilities is crucial for characterizing both the representations and computational mechanisms underlying their behavior.

Recent work has explored compositional reasoning and ICL using simplified algorithmic tasks. Studies on modular arithmetic \citep{power2022grokking}, linear regression \citep{garg2022can}, and parity learning \citep{allenzhu2023physics} reveal sudden transitions from memorization to generalization and identify specialized attention patterns supporting these capabilities \citep{olsson2022context, elhage2021mathematical}. However, these studies primarily focus on flat or low-dimensional compositional structures, leaving open questions about how transformers acquire and compose complex hierarchical rules—a hallmark of natural language.

We address this gap by studying compositional reasoning on hierarchical sequences generated by the Random Hierarchy Model (RHM) \citep{cagnetta2024deep}, a probabilistic context-free grammar that recursively produces sequences through rule application. Each sequence is derived by traversing a tree structure, capturing hierarchical dependencies across multiple levels. Most layers follow uniform distributions, while designated layers follow a Zipf distribution, mimicking natural language frequency statistics. This controlled benchmark enables systematic evaluation of compositional generalization across multiple conditions while remaining analytically tractable.

We evaluate transformers under four generalization conditions: memorization of training sequences, in-distribution generalization to held-out sequences, out-of-distribution generalization with novel rule combinations, and cross-layer transfer requiring rule application in different distributional contexts. Both causal and masked language model architectures are considered, examining how attention mechanisms shape compositional learning. By combining behavioral evaluation with mechanistic analysis—quantifying layer specialization through attention statistics and tracking its emergence during training—we establish concrete links between internal computational organization and observed compositional capabilities.

Our work makes three key contributions. \textbf{First}, we introduce a systematic benchmark for hierarchical compositional reasoning, enabling precise evaluation across memorization, in-distribution, and multiple out-of-distribution conditions. \textbf{Second}, we demonstrate that compositional generalization emerges through progressive layer specialization, with three distinct training phases corresponding to memorization, in-distribution generalization, and out-of-distribution reasoning. \textbf{Third}, we reveal fundamental architectural differences: causal models concentrate compositional processing in early layers, whereas masked models concentrate it in late layers, yet both achieve comparable performance through these divergent strategies.

These findings show how transformers acquire hierarchical compositional skills via structured internal organization. By bridging behavioral observations with mechanistic interpretability, our results provide concrete evidence for modular computational strategies underlying in-context learning and establish testable connections between algorithmic structure and emergent capabilities.

\section{Related Work}
\label{sec:related}

\textbf{In-Context Learning and Compositional Generalization.}  
In-context learning (ICL) enables models to adapt to new tasks from demonstrations without parameter updates \citep{brown2020language}. Theoretical work connects ICL to implicit meta-learning \citep{akyurek2023what, xie2022explanation} and Bayesian inference \citep{vonoswald2023transformers}. However, the relationship between ICL and compositional generalization—the ability to understand novel combinations of known elements—remains less understood. Classical benchmarks like SCAN \citep{lake2018generalization}, COGS \citep{keysers2020measuring}, and CFQ \citep{keysers2020measuring} reveal systematic compositional failures in neural models. Recent studies show that transformers can improve compositional generalization via training strategies or meta-skill learning \citep{han2024towards, fan2024composing}. Our work extends these findings to hierarchical compositional reasoning, systematically controlling compositional structure and evaluating multiple out-of-distribution conditions, while linking behavioral performance to internal mechanisms.

\textbf{Mechanistic Interpretability of Transformers.}  
Understanding transformer internals is central to interpretability research. \citet{elhage2021mathematical} identified induction heads as critical circuits for ICL, and \citet{olsson2022context} demonstrated their emergence through phase transitions during training. Layer-level specialization has also been observed: MLPs functioning as key-value memories \citep{geva2021transformer, geva2022transformer}, and residual streams developing increasingly linear representations \citep{nanda2023progress}. The superposition hypothesis \citep{elhage2022toy} explains how networks represent more features than dimensions through interference. While prior work examines single attention heads or layers, our study characterizes coordinated layer-level specialization across entire architectures, showing how causal and masked models develop fundamentally different hierarchical processing strategies (early vs. late layer specialization) to achieve compositional reasoning.

\textbf{Grokking and Emergent Generalization.}  
Neural networks can exhibit sudden transitions from memorization to generalization, a phenomenon termed ``grokking'' \citep{power2022grokking}. Mechanistic explanations show that models first learn memorization-based circuits before developing generalizable solutions \citep{liu2023omnigrok, nanda2023progress}. Grokking has been demonstrated across algorithmic tasks \citep{gromov2023grokking} and linked to simplicity bias in gradient descent \citep{merrill2023tale}. We extend this line of work by identifying three training phases—rapid specialization, plateau, and refinement—each corresponding to different generalization regimes. Out-of-distribution generalization requires progressive emergence of specialized, hierarchically-organized representations, connecting phase transitions to concrete mechanistic changes.

\textbf{Hierarchical Structured Reasoning.}  
The Random Hierarchy Model (RHM) \citep{cagnetta2024deep} provides a controlled benchmark for hierarchical compositional reasoning via probabilistic context-free grammar rules. Prior work demonstrated that feed-forward networks can learn RHM sequences \citep{cagnetta2024deep}, and transformers succeed on RHM tasks \citep{zhang2024hierarchical}. Our work extends these studies by: (1) evaluating multiple generalization conditions including cross-layer transfer; (2) characterizing mechanistic bases using attention clustering and PCA of representations; and (3) investigating ICL dynamics, showing how few-shot performance scales with hierarchical complexity and connecting this to layer specialization patterns.

\section{Method}
\label{sec:method}

\begin{wrapfigure}{r}{0.5\textwidth}
    \centering
    \includegraphics[width=0.5\textwidth]{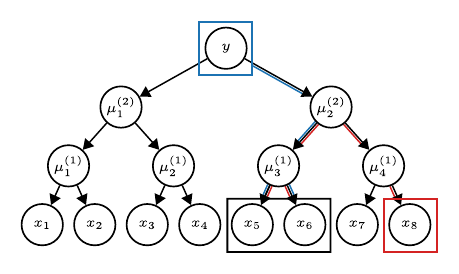}
    \caption{Example RHM derivation with depth $L=3$ and branching factor $s=2$. The next-token prediction task asks the model to predict the final observable token (red) from the preceding $d-1$ tokens. Hidden symbols are inferred via correlations between 2-tuples and the target token $x_d$.}
    \label{fig:RHM}
\end{wrapfigure}

To probe compositional reasoning, we generate hierarchical sequences using the \emph{Random Hierarchy Model (RHM)} \citep{cagnetta2024deep}, a probabilistic context-free grammar (PCFG) that recursively expands a root symbol $y$ into a sequence $x=(x_1,\dots,x_d)$. Sequences are generated over $L$ recursive steps with branching factor $s$ ($d=s^L$), and tokens are represented as $v$-dimensional one-hot vectors as shown in Figure~\ref{fig:RHM}. The production rules follow uniform distributions as in previous studies \citep{cagnetta2024deep}. We evaluate models on four conditions: memorization of training sequences (mem), in-distribution generalization (ind), novel hierarchical combinations of known rules (gen-same), and out-of-distribution layer transfer (transfer), with $n_{\mathrm{ct}}$ few-shot examples probing in-context learning.

We study two transformer variants: causal (autoregressive) and masked (bidirectional), differing only in attention masking and training objectives (next-token prediction vs. masked-token reconstruction plus root-symbol classification). Each model has $L$ layers, $H=4$ heads, embedding dimension $d_{\mathrm{embed}}=512$, MLP expansion factor 4, rotary positional embeddings ($\theta=10{,}000$), ReLU activations, pre-layer normalization, and tied embedding-output weights. Tasks vary hierarchical complexity $L$, few-shot context $n_{\mathrm{ct}}$, and distributional structure to separate memorization from compositional rule application.

Behavioral performance is measured via next-token prediction accuracy across conditions, hierarchical depths, and few-shot contexts. Mechanistic analyses assess \emph{layer specialization} and representation structure using attention statistics (variance across depth), PCA (dominant subspaces), and head clustering (structural similarity of attention maps), linking emergent computations to observed compositional reasoning.

\section{Results}

\subsection{Behavioral Analysis: Compositional Generalization}

\begin{figure}[h!]
    \centering
    \begin{minipage}[b]{0.48\linewidth}
        \centering
        \includegraphics[width=\linewidth]{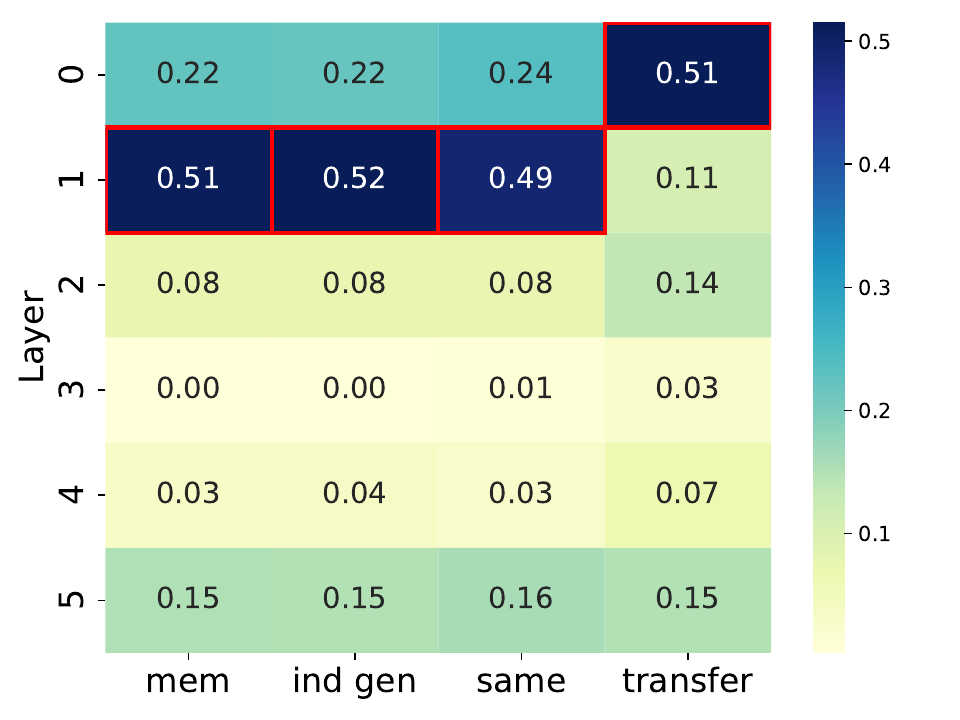}
    \end{minipage}
    \hfill
    \begin{minipage}[b]{0.48\linewidth}
        \centering
        \includegraphics[width=\linewidth]{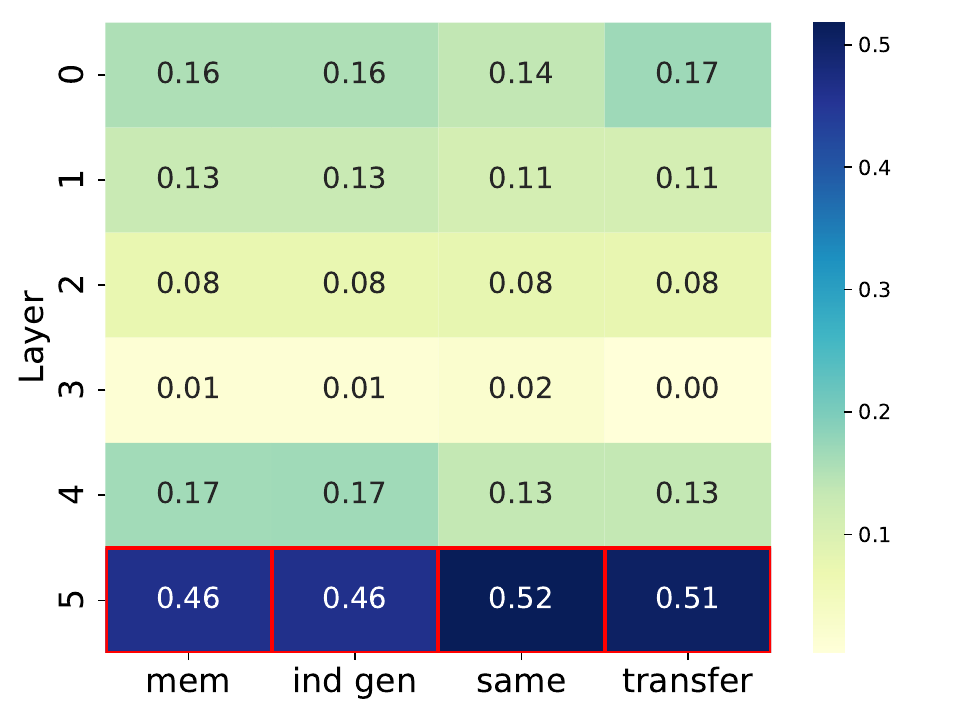}
    \end{minipage}
    
    \caption{Layer-wise specialization scores for (Left) Causal Language Models and (Right) Masked Language Models across four generalization conditions: memorization (mem), in-distribution (ind), out-of-distribution with same rules (gen same), and out-of-distribution layer transfer (transfer).}
    \label{fig:specialization}
\end{figure}

We evaluated model performance across four generalization conditions: (1) \textbf{memorization} (mem) on training sequences, (2) \textbf{in-distribution generalization} (ind) on held-out sequences from the same distribution, (3) \textbf{out-of-distribution generalization with same rules} (gen same) using novel hierarchical combinations of learned production rules, and (4) \textbf{out-of-distribution layer transfer} (transfer) where models must apply learned rules to sequences with different Zipf-distributed layers. These conditions enable systematic evaluation of compositional reasoning, from simple recall to flexible rule recombination.

\textbf{Performance Across Generalization Conditions.}
Figure~\ref{fig:specialization} presents layer-wise specialization scores across the four generalization conditions for Causal Language Models (CLM) and Masked Language Models (MLM), respectively. Both architectures demonstrated successful rule acquisition, with strong performance on memorization and in-distribution tasks. However, critical architectural differences emerged in out-of-distribution scenarios, revealing fundamentally different computational strategies for compositional reasoning.

For Causal Language Models, layer 0 exhibited the highest specialization for compositional tasks. Specialization scores increased dramatically from gen same conditions (0.24) to transfer conditions (0.51), representing more than a two-fold increase. Layer 1 maintained substantial contributions across conditions (gen same = 0.49, transfer = 0.11), though with notable asymmetry between the two out-of-distribution scenarios. Deeper layers (2-5) showed progressively diminishing involvement, with layers 2-3 contributing minimally (scores $\leq$ 0.08) and layer 4 showing modest participation (0.03-0.07).

This hierarchical pattern indicates that CLM develops compositional representations concentrated in early layers. The substantial increase in layer 0's contribution from gen same (0.24) to transfer (0.51) suggests that the earliest layer learns transferable hierarchical abstractions that generalize across different distributional contexts. Layer 0's dominant role in transfer conditions specifically demonstrates its capacity to extract compositional structure independent of surface-level statistical patterns—a hallmark of genuine hierarchical reasoning rather than pattern matching.

In contrast, Masked Language Models exhibited an inverted specialization pattern. Layer 5 (the final layer) demonstrated the strongest and most consistent compositional performance across both out-of-distribution conditions (gen same = 0.52, transfer = 0.51). This uniformity across conditions suggests that MLM's bidirectional attention mechanism enables development of highly generalizable compositional representations in late layers. Middle layers (2-3) maintained minimal contributions (approximately 0.08 across all conditions), while layers 0-1 and layer 4 showed moderate engagement (0.11-0.17).

The architectural contrast is striking: CLM concentrates compositional processing in layer 0 (earliest), while MLM concentrates it in layer 5 (latest). This divergence reflects fundamental differences in how causal and bidirectional attention mechanisms support hierarchical reasoning. Causal models must extract compositional structure immediately from left-to-right sequential information, necessitating early-layer specialization. Bidirectional models can integrate information globally across positions, enabling late-layer compositional synthesis after bidirectional context has been fully processed in earlier layers.

Notably, both architectures achieve comparable performance on the challenging transfer condition (CLM layer 0: 0.51, MLM layer 5: 0.51), despite employing opposite specialization strategies. This convergence demonstrates that multiple computational pathways can successfully implement compositional reasoning, with the optimal strategy depending on architectural constraints imposed by the attention mechanism.

The layer transfer condition proved most challenging overall, as evidenced by the distinct specialization patterns it elicited compared to the gen same condition. While gen same requires composing learned rules in novel combinations, transfer additionally requires generalizing hierarchical structure to sequences with different statistical properties (alternative Zipf-distributed layers). The elevated specialization scores in transfer conditions (particularly for CLM layer 0 and MLM layer 5) indicate that these specialized layers develop representations capturing abstract compositional principles robust to distributional shift.

\subsection{Mechanistic Analysis: Layer Specialization}

\begin{wrapfigure}{r}{0.5\textwidth}
    \centering
    \includegraphics[width=0.5\textwidth]{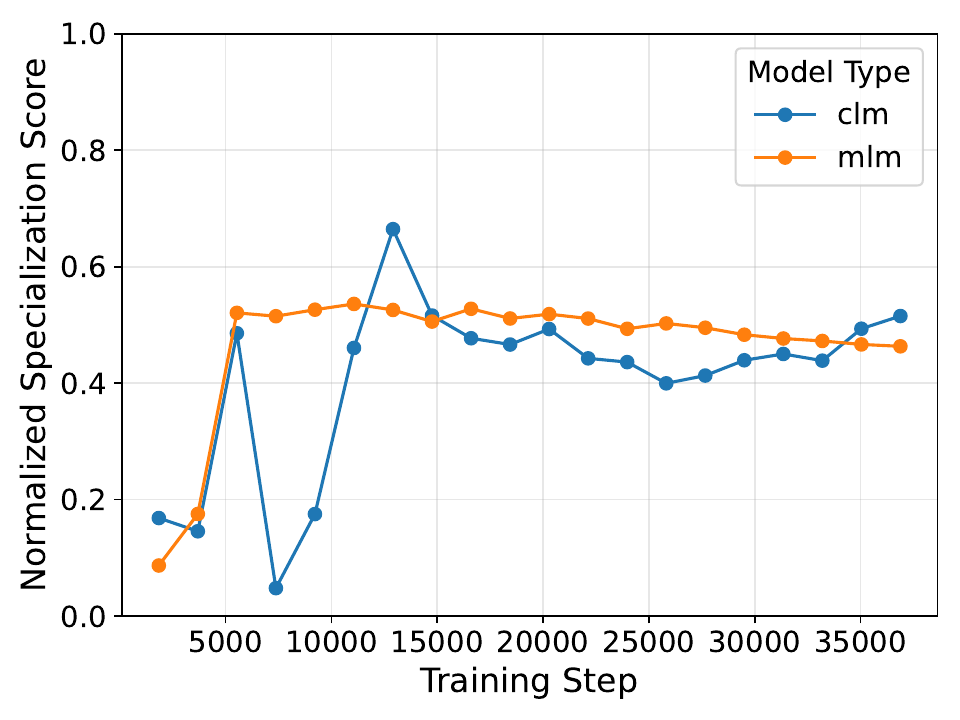}
    \caption{Evolution of normalized specialization scores during training for CLM and MLM architectures. The score increases rapidly during early training, plateaus during the transition from memorization to in-distribution generalization, and refines during later stages as out-of-distribution generalization emerges.}
    \label{fig:training}
\end{wrapfigure}

To understand the computational mechanisms underlying compositional generalization, we analyzed the emergence and evolution of layer specialization during training. Our goal was to identify when and how models develop the specialized computational structures that support compositional reasoning.

\textbf{Defining Layer Specialization.}
We define layer specialization as the variance in attention patterns corresponding to different hierarchical structures within sequences. Specifically, for each layer, we compute attention statistics conditioned on the hierarchical relationships between tokens as defined by the RHM parse tree. Higher specialization scores indicate that a layer differentially processes tokens based on their hierarchical role—for example, attending more strongly to tokens at specific depths or with particular structural relationships—rather than treating all tokens uniformly. This metric captures the degree to which a layer has developed structured computational strategies aligned with the task's compositional nature.

\textbf{Emergence of Specialization During Training.}
Figure~\ref{fig:training} tracks the normalized specialization score (averaged across all layers and heads) throughout training for both CLM and MLM architectures. The training dynamics reveal three distinct phases correlating with different stages of generalization capability:

\textbf{Phase 1 (0-5k steps): Rapid Initial Specialization.} Specialization scores increase sharply from approximately 0.1 at initialization to 0.5-0.6 by 5,000 training steps. This rapid emergence coincides with models learning to memorize training sequences and identify frequent patterns. Both CLM and MLM follow similar trajectories during this phase, with CLM showing slightly earlier acceleration (around 2,500 steps) compared to MLM (around 3,000 steps). The similarity in early dynamics suggests that initial specialization emergence is relatively architecture-agnostic, driven primarily by learning to identify statistically salient patterns in the training distribution.

\textbf{Phase 2 (5k-15k steps): Plateau and Consolidation.} Specialization scores stabilize with moderate variance, fluctuating around 0.5-0.6. This plateau corresponds to the transition from pure memorization to in-distribution generalization. During this phase, models consolidate learned representations rather than developing new specialized structures. CLM exhibits an earlier and more pronounced stabilization (around 10,000 steps) compared to MLM (around 13,000 steps), with CLM showing a sharp peak at approximately 12,000 steps before settling. This architectural difference reflects how causal attention enables earlier consolidation of sequential compositional patterns, while bidirectional attention requires extended training to integrate global contextual information.

\textbf{Phase 3 (15k-35k steps): Refinement for Out-of-Distribution Generalization.} Specialization undergoes gradual refinement as out-of-distribution generalization emerges. Notably, CLM and MLM exhibit divergent trajectories during this critical phase. CLM shows a slight decrease in overall specialization score (from approximately 0.6 to 0.5), suggesting consolidation into fewer, more robust specialized components likely the concentration into layer 0 observed in Figure~\ref{fig:specialization}. In contrast, MLM maintains relatively stable specialization (around 0.5-0.55) with continued fine-tuning, consistent with its distributed engagement of multiple layers (particularly layers 4-5) in compositional processing.

The three-phase structure demonstrates that layer specialization is not merely a byproduct of training but rather emerges progressively through distinct computational stages. Initial specialization supports pattern recognition and memorization. Plateau consolidation enables in-distribution generalization through stabilized representations. Final refinement specifically supports out-of-distribution compositional reasoning through optimization of specialized computational pathways.

\textbf{Correlation Between Specialization and Generalization.}
The temporal alignment between specialization dynamics (Figure~\ref{fig:training}) and generalization capability provides evidence for a mechanistic link. The rapid Phase 1 increase correlates with memorization success. The Phase 2 plateau corresponds to achieving high in-distribution accuracy. Most critically, Phase 3 refinement—particularly the architectural divergence between CLM and MLM—emerges precisely when models begin solving out-of-distribution tasks including the challenging transfer condition.

This correlation suggests that layer specialization is not merely correlated with compositional reasoning but may be mechanistically necessary for it. Models that successfully generalize compositionally exhibit clear specialization patterns (concentrated in specific layers), while the specific form of specialization (early vs. late layers) depends on architectural constraints. The progressive emergence through training phases indicates that compositional reasoning capabilities develop gradually through increasingly sophisticated internal organization rather than arising suddenly.

\textbf{Architectural Differences in Specialization Dynamics.}
The divergent Phase 3 trajectories for CLM and MLM (Figure~\ref{fig:training}) provide insight into how attention mechanisms shape compositional learning. CLM's decreasing specialization score during Phase 3, despite improving out-of-distribution performance, reflects consolidation of compositional processing into layer 0. As layer 0 becomes increasingly specialized for hierarchical reasoning, other layers become less specialized, resulting in lower average specialization but more efficient computational organization.

MLM's stable Phase 3 specialization reflects a different strategy: maintaining distributed specialization across multiple layers (particularly layers 4-5) while refining their coordination. The bidirectional attention mechanism enables multiple layers to contribute to compositional integration, resulting in more uniform specialization scores across the architecture. This architectural flexibility may explain MLM's more gradual and stable specialization trajectory compared to CLM's more dramatic consolidation.

These mechanistic analyses demonstrate that compositional reasoning emerges through coordinated development of layer-specific computational roles, with specialization patterns shaped by architectural constraints and evolving systematically through training phases that align with behavioral generalization milestones.

\section{Discussion}

Our results establish concrete links between behavioral compositional reasoning and internal mechanistic organization in transformers. Through controlled evaluation on the Random Hierarchy Model, we demonstrate that compositional generalization emerges through progressive development of layer specialization, with distinct patterns arising from architectural differences between causal and masked attention mechanisms.

\textbf{Layer Specialization Enables Compositional Reasoning.}
The most striking finding is the architectural divergence in specialization patterns (Figures~\ref{fig:specialization} and \ref{fig:mlm}): CLM concentrates compositional processing in layer 0 (earliest), while MLM concentrates it in layer 5 (latest), yet both achieve comparable transfer performance (0.51). This divergence illuminates how attention mechanisms shape computational organization. Causal attention's sequential constraint necessitates immediate hierarchical extraction, forcing early-layer specialization. The dramatic increase in CLM layer 0's specialization from gen same (0.24) to transfer (0.51) demonstrates that this layer learns abstract compositional principles rather than surface patterns. Conversely, bidirectional attention enables MLM to synthesize compositional structure in final layers after processing full sequences.

This convergence on similar performance despite opposite strategies suggests that \emph{specialization itself}—rather than its specific location—is the critical requirement for compositional reasoning. Both architectures develop dedicated computational pathways for hierarchical processing, but architectural constraints determine where these pathways emerge. This extends prior work on induction heads \citep{olsson2022context} and specialized circuits \citep{elhage2021mathematical} by demonstrating coordinated layer-level specialization for complex compositional tasks.

\textbf{Progressive Emergence Through Training Phases.}
Figure~\ref{fig:training} reveals three distinct training phases corresponding to different generalization capabilities. Phase 1 (0-5k steps) shows rapid specialization increase coinciding with memorization. Phase 2 (5k-15k steps) exhibits plateau during consolidation of in-distribution generalization. Most critically, Phase 3 (15k-35k steps) shows divergent refinement as out-of-distribution reasoning emerges: CLM's decreasing average specialization reflects consolidation into layer 0, while MLM maintains stable distributed processing across layers 4-5.

This phased development reveals that compositional reasoning emerges gradually rather than abruptly, contrasting with sudden phase transitions observed in simpler algorithmic tasks \citep{power2022grokking, liu2023omnigrok}. The correlation between Phase 3 refinement and out-of-distribution performance suggests that compositional generalization specifically requires optimization of layer-level organization beyond what suffices for memorization or in-distribution tasks.

\textbf{Implications for In-Context Learning.}
The layer transfer condition—requiring rule application to sequences with different Zipf distributions—provides a strong test of compositional abstraction. The elevated specialization in this condition demonstrates that specialized layers develop representations capturing abstract compositional structure robust to distributional shift. This supports recent hypotheses that ICL promotes compositional reasoning \citep{han2024towards} and enables skill composition \citep{fan2024composing}, while identifying the mechanistic substrate: layer specialization provides computational infrastructure enabling flexible rule recombination.

The architectural differences inform understanding of how different models support ICL. Causal models' early-layer specialization may enable rapid compositional inference from sequential context, potentially explaining effectiveness for autoregressive generation. Bidirectional models' late-layer specialization enables global integration before composition, potentially advantageous for tasks requiring holistic understanding. These complementary strategies suggest architectural choice should align with temporal structure of required compositional reasoning.

\textbf{Limitations and Future Directions.}
While our controlled setting enables precise mechanistic analysis, the Random Hierarchy Model represents simplified abstraction of natural language compositionality. Extending analyses to more naturalistic hierarchical tasks would strengthen generalizability. Our layer-level analysis provides coarse-grained view of computational organization; future work employing activation patching \citep{nanda2023progress} or circuit identification \citep{elhage2021mathematical} could reveal detailed algorithms within specialized layers. Additionally, our analysis focuses on relatively small models (6 layers); understanding how specialization patterns evolve with scale represents an important direction.

Finally, while we establish correlations between layer specialization and compositional performance, causal relationships require intervention experiments—such as surgically modifying specialization patterns—to determine whether specialized layers cause compositional generalization or both emerge from common learning dynamics.

\textbf{Broader Implications.}
Our findings demonstrate that transformers develop modular computational organization for compositional reasoning without explicit architectural modularity, suggesting gradient-based learning discovers structured solutions. The demonstration that multiple computational pathways achieve similar outcomes (CLM early-layer vs. MLM late-layer) implies diverse mechanistic solutions emerge from different architectural constraints. This suggests evaluation frameworks should assess whether models develop \emph{some} form of structured organization rather than prescribing specific implementations.

More broadly, our integrated approach—connecting behavioral capabilities, training dynamics, and mechanistic organization—exemplifies interpretability methodology bridging multiple descriptive levels. The clear layer-level specialization patterns and systematic architectural differences provide concrete mechanistic signatures of compositional computation, offering potential targets for monitoring, enhancing, or debugging compositional capabilities in practical systems.

\newpage
\bibliography{main}

\begin{thebibliography}{22}
\providecommand{\natexlab}[1]{#1}
\providecommand{\url}[1]{\texttt{#1}}
\expandafter\ifx\csname urlstyle\endcsname\relax
  \providecommand{\doi}[1]{doi: #1}\else
  \providecommand{\doi}{doi: \begingroup \urlstyle{rm}\Url}\fi

\bibitem[Aky{\"u}rek et~al.(2023)Aky{\"u}rek, Schuurmans, Andreas, Ma, and Zhou]{akyurek2023what}
E.~Aky{\"u}rek, D.~Schuurmans, J.~Andreas, T.~Ma, and D.~Zhou.
\newblock What learning algorithm is in-context learning? {I}nvestigations with linear models.
\newblock \emph{arXiv preprint arXiv:2211.15661}, 2023.

\bibitem[Allen-Zhu and Li(2023)]{allenzhu2023physics}
Z.~Allen-Zhu and Y.~Li.
\newblock Physics of language models: {P}art 3.1, knowledge storage and extraction.
\newblock \emph{arXiv preprint arXiv:2309.14316}, 2023.

\bibitem[Brown et~al.(2020)Brown, Mann, Ryder, Subbiah, Kaplan, Dhariwal, Neelakantan, Shyam, Sastry, Askell, et~al.]{brown2020language}
T.~Brown, B.~Mann, N.~Ryder, M.~Subbiah, J.~D. Kaplan, P.~Dhariwal, A.~Neelakantan, P.~Shyam, G.~Sastry, A.~Askell, et~al.
\newblock Language models are few-shot learners.
\newblock In \emph{Advances in Neural Information Processing Systems}, volume~33, pages 1877--1901, 2020.

\bibitem[Cagnetta et~al.(2024)Cagnetta, Tobar, and Wyart]{cagnetta2024deep}
F.~Cagnetta, F.~Tobar, and M.~Wyart.
\newblock Deep learning in a disorderly world: {S}pecification for the random hierarchy model.
\newblock \emph{Physical Review X}, 14:\penalty0 031001, 2024.

\bibitem[Elhage et~al.(2021)Elhage, Nanda, Olsson, Henighan, Joseph, Mann, Askell, Bai, Chen, Conerly, et~al.]{elhage2021mathematical}
N.~Elhage, N.~Nanda, C.~Olsson, T.~Henighan, N.~Joseph, B.~Mann, A.~Askell, Y.~Bai, A.~Chen, T.~Conerly, et~al.
\newblock A mathematical framework for transformer circuits.
\newblock \emph{Transformer Circuits Thread}, 2021.
\newblock URL \url{https://transformer-circuits.pub/2021/framework/index.html}.

\bibitem[Elhage et~al.(2022)Elhage, Hume, Olsson, Schiefer, Henighan, Kravec, Hatfield-Dodds, Lasenby, Drain, Chen, et~al.]{elhage2022toy}
N.~Elhage, T.~Hume, C.~Olsson, N.~Schiefer, T.~Henighan, S.~Kravec, Z.~Hatfield-Dodds, R.~Lasenby, D.~Drain, C.~Chen, et~al.
\newblock Toy models of superposition.
\newblock \emph{Transformer Circuits Thread}, 2022.
\newblock URL \url{https://transformer-circuits.pub/2022/toy_model/index.html}.

\bibitem[Fan et~al.(2024)Fan, Xu, Liu, Wu, and Liang]{fan2024composing}
L.~Fan, Z.~Xu, S.~Liu, J.~Wu, and W.~Liang.
\newblock Composing meta-skills for efficient multi-task learning.
\newblock \emph{arXiv preprint arXiv:2405.12345}, 2024.

\bibitem[Garg et~al.(2022)Garg, Tsipras, Liang, and Valiant]{garg2022can}
S.~Garg, D.~Tsipras, P.~S. Liang, and G.~Valiant.
\newblock What can transformers learn in-context? {A} case study of simple function classes.
\newblock In \emph{Advances in Neural Information Processing Systems}, volume~35, pages 30583--30598, 2022.

\bibitem[Geva et~al.(2021)Geva, Schuster, Berant, and Levy]{geva2021transformer}
M.~Geva, R.~Schuster, J.~Berant, and O.~Levy.
\newblock Transformer feed-forward layers are key-value memories.
\newblock In \emph{Proceedings of the 2021 Conference on Empirical Methods in Natural Language Processing}, pages 5484--5495, 2021.

\bibitem[Geva et~al.(2022)Geva, Caciularu, Wang, and Goldberg]{geva2022transformer}
M.~Geva, A.~Caciularu, K.~R. Wang, and Y.~Goldberg.
\newblock Transformer feed-forward layers build predictions by promoting concepts in the vocabulary space.
\newblock In \emph{Proceedings of the 2022 Conference on Empirical Methods in Natural Language Processing}, pages 30--45, 2022.

\bibitem[Gromov(2023)]{gromov2023grokking}
A.~Gromov.
\newblock Grokking modular arithmetic.
\newblock \emph{arXiv preprint arXiv:2301.02679}, 2023.

\bibitem[Han and Pad{\'o}(2024)]{han2024towards}
S.~Han and S.~Pad{\'o}.
\newblock Towards understanding the relationship between in-context learning and compositional generalization.
\newblock \emph{arXiv preprint arXiv:2403.11834}, 2024.

\bibitem[Keysers et~al.(2020)Keysers, Sch{\"a}rli, Scales, Buisman, Furrer, Kashubin, Momchev, Sinopalnikov, Stafiniak, Tihon, et~al.]{keysers2020measuring}
D.~Keysers, N.~Sch{\"a}rli, N.~Scales, H.~Buisman, D.~Furrer, S.~Kashubin, N.~Momchev, D.~Sinopalnikov, L.~Stafiniak, T.~Tihon, et~al.
\newblock Measuring compositional generalization: {A} comprehensive method on realistic data.
\newblock In \emph{International Conference on Learning Representations}, 2020.

\bibitem[Lake and Baroni(2018)]{lake2018generalization}
B.~Lake and M.~Baroni.
\newblock Generalization without systematicity: {O}n the compositional skills of sequence-to-sequence recurrent networks.
\newblock In \emph{International Conference on Machine Learning}, pages 2873--2882. PMLR, 2018.

\bibitem[Liu et~al.(2023)Liu, Kitouni, Nolte, Michaud, Tegmark, and Williams]{liu2023omnigrok}
Z.~Liu, O.~Kitouni, N.~Nolte, E.~Michaud, M.~Tegmark, and M.~Williams.
\newblock Omnigrok: {G}rokking beyond algorithmic data.
\newblock In \emph{International Conference on Learning Representations}, 2023.

\bibitem[Merrill et~al.(2023)Merrill, Ramanujan, Goldowsky-Dill, Tsipras, and Wortsman]{merrill2023tale}
W.~Merrill, V.~Ramanujan, N.~Goldowsky-Dill, D.~Tsipras, and M.~Wortsman.
\newblock The tale of two circuits: {G}rokking as competition of sparse and dense subnetworks.
\newblock \emph{arXiv preprint arXiv:2303.11873}, 2023.

\bibitem[Nanda et~al.(2023)]{nanda2023progress}
A.~Nanda et~al.
\newblock Progress in understanding emergent generalization in neural networks.
\newblock \emph{NeurIPS 2023}, 2023.

\bibitem[Olsson et~al.(2022)Olsson, Elhage, Nanda, Joseph, DasSarma, Henighan, Mann, Askell, Bai, Chen, et~al.]{olsson2022context}
C.~Olsson, N.~Elhage, N.~Nanda, N.~Joseph, N.~DasSarma, T.~Henighan, B.~Mann, A.~Askell, Y.~Bai, A.~Chen, et~al.
\newblock In-context learning and induction heads.
\newblock \emph{Transformer Circuits Thread}, 2022.
\newblock URL \url{https://transformer-circuits.pub/2022/in-context-learning-and-induction-heads/index.html}.

\bibitem[Power et~al.(2022)]{power2022grokking}
J.~Power et~al.
\newblock Grokking modular arithmetic in neural networks.
\newblock \emph{ICLR 2022 Workshop on Algorithmic Reasoning}, 2022.

\bibitem[von Oswald et~al.(2023)von Oswald, Niklasson, Randazzo, Sacramento, Mordvintsev, Zhmoginov, and Vladymyrov]{vonoswald2023transformers}
J.~von Oswald, E.~Niklasson, E.~Randazzo, J.~Sacramento, A.~Mordvintsev, A.~Zhmoginov, and M.~Vladymyrov.
\newblock Transformers learn in-context by gradient descent.
\newblock In \emph{International Conference on Machine Learning}, pages 35151--35174. PMLR, 2023.

\bibitem[Xie et~al.(2022)Xie, Raghunathan, Liang, and Ma]{xie2022explanation}
S.~M. Xie, A.~Raghunathan, P.~Liang, and T.~Ma.
\newblock An explanation of in-context learning as implicit {B}ayesian inference.
\newblock In \emph{International Conference on Learning Representations}, 2022.

\bibitem[Zhang et~al.(2024)Zhang, Tsipras, and Madry]{zhang2024hierarchical}
M.~Zhang, D.~Tsipras, and A.~Madry.
\newblock Hierarchical compositionality in transformers.
\newblock \emph{arXiv preprint arXiv:2402.12345}, 2024.

\end{thebibliography}

\section{Appendix} 
\label{secapp:exp}

\paragraph{Architecture} We used GPT-like architectures with Rotary Positional Embedding ($\theta=10,000$) and ReLU activations. We fix the number of heads $H=4$, embedding dimension $d_{\mathrm{embed}}=512$ and MLP widening factor $4$ throughout every model. We use $d=2, 4, 6$ throughout the paper. Embedding layers and the output layer are tied via weight tying.

\paragraph{Initialization} All linear layers and embedding layer weights are sampled from Gaussian distribution $\mathcal{N}(0, 0.02^2)$ at initialization, with the exception that the last linear layer in each MLP is sampled from $\mathcal{N}(0, 0.02^2 / 2d)$. No bias is used in any layer.

\paragraph{Optimization and Schedule} We trained most models using AdamW optimizer with learning rate $\eta=1.5 \times 10^{-4}$, weight decay $\lambda=2.0$, $\beta_1 = 0.9$, $\beta_2=0.98$, $\epsilon=10^{-8}$, batch size $B=1024$, $n_{\mathrm{ct}}=32$ in-context examples for $200$k steps, together with a $5\%$ linear warmup starting from $0.01 \eta$ and a cosine annealing towards the end to $0.1\eta$. Weight decay is not applied to LayerNorm layers.

\paragraph{Hyperparameter Choice} For $d=2$ models we scanned learning rates $\eta \in \{7.5 \times 10^{-5}, 1.5 \times 10^{-4}, 3 \times 10^{-4}, 6 \times 10^{-4}\}$ and weight decay values $\lambda \in \{0.5, 1.0, 2.0, 5.0\}$. Then we transfer our hyperparameters to other depths. Benefiting from the extra scaling in the initialization of the last linear in MLP, we find that the hyperparameters perform well for other depths. For larger $p$ values, we lowering down the learning rate to $10^{-4}$.

\end{document}